\documentclass[runningheads]{llncs}
\usepackage{graphicx}
\usepackage{amsmath,amssymb} 
\usepackage{color}
\usepackage[utf8]{inputenc}
\usepackage[english]{babel}
\usepackage{hyperref}
\usepackage{subcaption}
\usepackage[capitalise]{cleveref}
\usepackage{booktabs}
\usepackage{tabularx}
\usepackage[table]{xcolor}

\begin{document}

\newcommand{\point}{
    \raise0.7ex\hbox{.}
    }

\pagestyle{headings}

\mainmatter

\title{Semantic Segmentation of Earth Observation Data Using Multimodal and Multi-scale Deep Networks}
\titlerunning{Semantic Segmentation of EO Data Using Deep Nets}
\authorrunning{Nicolas Audebert, Bertrand Le Saux, Sébastien Lefèvre}

\author{Nicolas Audebert\textsuperscript{1,2}, Bertrand Le Saux\textsuperscript{1}, Sébastien Lefèvre\textsuperscript{2}}
\institute{\textsuperscript{1 }ONERA, \textit{The French Aerospace Lab}, F-91761 Palaiseau, France - \texttt{\{nicolas.audebert,bertrand.le\_saux\}@onera.fr}\\
	\textsuperscript{2 }Univ. Bretagne-Sud, UMR 6074, IRISA, F-56000 Vannes, France - \texttt{sebastien.lefevre@irisa.fr}\\}

\maketitle

\newcolumntype{Y}{>{\centering\arraybackslash}X}

\newsavebox{\dualsegnet}

\begin{abstract}
This work investigates the use of deep fully convolutional neural networks (DFCNN) for pixel-wise scene labeling of Earth Observation images. Especially, we train a variant of the SegNet architecture on remote sensing data over an urban area and study different strategies for performing accurate semantic segmentation. Our contributions are the following: 1)~we transfer efficiently a DFCNN from generic everyday images to remote sensing images; 2)~we introduce a multi-kernel convolutional layer for fast aggregation of predictions at multiple scales; 3)~we perform data fusion from heterogeneous sensors (optical and laser) using residual correction. Our framework improves state-of-the-art accuracy on the ISPRS Vaihingen 2D Semantic Labeling dataset.
\end{abstract}

\section{Introduction}

Over the past few years, deep learning has become ubiquitous for computer vision tasks. Convolutional Neural Networks (CNN) took over the field and are now the state-of-the-art for object classification and detection. Recently, deep networks extended their abilities to semantic segmentation, thanks to recent works designing deep networks for dense (pixel-wise) prediction, generally built around the fully convolutional principle stated by Long et al.~\cite{long_fully_2015}. These architectures have gained a lot of interest during the last years thanks to their ability to address semantic segmentation. Indeed, fully convolutional architectures are now considered as the state-of-the-art on most renowned benchmarks such as PASCAL VOC2012~\cite{everingham_pascal_2014} and Microsoft COCO~\cite{lin_microsoft_2014}. However, those datasets focus on everyday scenes and assume a human-level point of view. In this work, we aim to process remote sensing (RS) data and more precisely Earth Observation (EO) data. EO requires to extract thematic information (e.g. land cover usage, biomass repartition, etc.) using data acquired from various airborne and/or satellite sensors (e.g. optical cameras, LiDAR). It often relies on a mapping step, that aims to automatically produce a semantic map containing various regions of interest, based on some raw data. A popular application is land cover mapping where each pixel is assigned to a thematic class, according to the type of land cover (vegetation, road, \dots) or object (car, building, \dots) observed at the pixel coordinates. As volume of EO data continuously grows (reaching the Zettabyte scale), deep networks can be trained to understand those images. However, there are several strong differences between everyday pictures and EO imagery. First, EO assumes a bird's view acquisition, thus the perspective is significantly altered w.r.t. usual computer vision datasets. Objects lie within a flat 2D plane, which makes the angle of view consistent but reduces the number of depth-related hints, such as projected shadows. Second, every pixel in RS images has a semantic meaning. This differs from most images in the PASCAL VOC2012 dataset, that are mainly comprised of a meaningless background with a few foreground objects of interest. Such a distinction is not as clear in EO data, where images may contain both semantically meaningful ``stuff'' (large homogeneous non quantifiable surfaces such as water bodies, roads, corn fields, \dots) and ``objects'' (cars, houses, \dots) that have different properties.

First experiments using deep learning introduced CNN for classification of EO data with a patch based approach~\cite{lagrange_benchmarking_2015}. Images were segmented using a segmentation algorithm (e.g. with superpixels) and each region was classified using a CNN. However, the unsupervised segmentation proved to be a difficult bottleneck to overcome as higher accuracy requires strong oversegmentation. This was improved thanks to CNN using dense feature maps~\cite{paisitkriangkrai_effective_2015}. Fully supervised learning of both segmentation and classification is a promising alternative that could drastically improve the performance of the deep models. Fully convolutional networks~\cite{long_fully_2015} and derived models can help solve this problem. Adapting these architectures to multimodal EO data is the main objective of this work.

In this work, we show how to perform competitive semantic segmentation of EO data. We consider a standard dataset delivered by the ISPRS~\cite{rottensteiner_isprs_2012} and rely on deep fully convolutional networks, designed for dense pixel-wise prediction. Moreover, we build on this baseline approach and present a simple trick to smooth the predictions using a multi-kernel convolutional layer that operates several parallel convolutions with different kernel sizes to aggregate predictions at multiple scale. This module does not need to be retrained from scratch and smoothes the predictions by averaging over an ensemble of models considering multiple scales, and therefore multiple spatial contexts. Finally, we present a data fusion method able to integrate auxiliary data into the model and to merge predictions using all available data. Using a dual-stream architecture, we first naively average the predictions from complementary data. Then, we introduce a residual correction network that is able to learn how to fuse the prediction maps by adding a corrective term to the average prediction.

\section{Related Work}

\subsection{Semantic Segmentation}
In computer vision, semantic segmentation consists in assigning a semantic label (i.e. a class) to each coherent region of an image. This can be achieved using pixel-wise dense prediction models that are able to classify each pixel of the image. Recently, deep learning models for semantic segmentation have started to appear. Many recent works in computer vision are actually tackling semantic segmentation with a significant success. Nearly all state-of-the-art architectures follow principles stated in~\cite{long_fully_2015}, where semantic segmentation using Fully Convolutional Networks (FCN) has been shown to achieve impressive results on PASCAL VOC2012. The main idea consists in modifying traditional classification CNN so that the output is not a probability vector but rather a probability map. Generally, a standard CNN is used as an encoder that will extract features, followed by a decoder that will upsample feature maps to the original spatial resolution of the input image. A heat map is then obtained for each class. Following the path opened by FCN, several architectures have proven to be very effective on both PASCAL VOC2012 and Microsoft COCO. Progresses have been obtained by increasing the field-of-view of the encoder and removing pooling layers to avoid bottlenecks (DeepLab~\cite{chen_semantic_2015} and dilated convolutions~\cite{yu_multi-scale_2015}). Structured prediction has been investigated with integrated structured models such as Conditional Random Fields (CRF) within the deep network (CRFasRNN~\cite{zheng_conditional_2015,arnab_higher_2015}). Better architectures also provided new insights (e.g. ResNet~\cite{he_deep_2016} based architectures~\cite{wu_high-performance_2016}, recurrent neural networks~\cite{yan_combining_2016}). Leveraging analogies with convolutional autoencoders (and similarly to Stacked What-Where Autoencoders~\cite{zhao_stacked_2015}), DeconvNet~\cite{noh_learning_2015} and SegNet~\cite{badrinarayanan_segnet:_2015} have investigated symmetrical encoder-decoder architectures.
  
\subsection{Scene Understanding in Earth Observation Imagery}
  
Deep learning on EO images is a very active research field. Since the first works on road detection~\cite{mnih_learning_2010}, CNN have been successfully used for classification and dense labeling of EO data. CNN-based deep features have been shown to outperform significantly traditional methods based on hand-crafted features and Support Vector Machines for land cover classification~\cite{penatti_deep_2015}. Besides, a framework using superpixels and deep features for semantic segmentation outperformed traditional methods~\cite{lagrange_benchmarking_2015} and obtained a very high accuracy in the Data Fusion Contest~2015~\cite{campos-taberner_processing_2016}. A generic deep learning framework for processing remote sensing data using CNN established that deep networks improve significantly the commonly used SVM baseline~\cite{nogueira_towards_2016}. \cite{zhao_learning_2016}~also performed classification of EO data using ensemble of multiscale CNN, which has been improved with the introduction of FCN~\cite{marmanis_semantic_2016}. Indeed, fully convolutional architectures are promising as they can learn how to classify the pixels (``what'') but also predict spatial structures (``where''). Therefore, on EO images, such models would be not only able to detect different types of land cover in a patch, but also to predict the shapes of the buildings, the curves of the roads, \dots

\section{Proposed Method}

\subsection{Data Preprocessing}
\label{sec:data_proc}

High resolution EO images are often too large to be processed in only one pass through a CNN. For example, the average dimensions of an ISPRS tile from Vaihingen dataset is $2493 \times 2063$ pixels, whereas most CNN are tailored for a resolution of $256 \times 256$ pixels. Given current GPU memory limitations, we split our EO images in smaller patches with a simple sliding window. It is then possible to process arbitrary large images in a linear time. In the case where consecutive patches overlap at testing time (if the stride is smaller than the patch size), we average the multiple predictions to obtain the final classification for overlapping pixels. This smoothes the predictions along the borders of each patch and removes the discontinuities that can appear.

We recall that our aim is to transpose well-known architectures from traditional computer vision to EO. We are thus using neural networks initially designed for RGB data. Therefore, the processed images will have to respect such a 3-channel format. The ISPRS dataset contains IRRG images of Vaihingen. The 3 channels (i.e. near-infrared, red and green) will thus be processed as an RGB image. Indeed, all three color channels have been acquired by the same sensor and are the consequence of the same physical phenomenon. These channels have homogeneous dynamics and meaning for our remote sensing application. The dataset also includes additional data acquired from an aerial laser sensor and consisting of a Digital Surface Model (DSM). In addition, we also use the Normalized Digital Surface Model (NDSM) from~\cite{gerke_use_2015}. Finally, we compute the Normalized Difference Vegetation Index (NDVI) from the near-infrared and red channels. NDVI is a good indicator for vegetation and is computed as follows:
\begin{equation}
NDVI = {\frac{IR - R}{IR + R}}~.
\end{equation}

Let us recall that we are working in a 3-channel framework. Thus we build for each IRRG image another companion composite image using the DSM, NDSM and NDVI information.
Of course, such information does not correspond to color channels and cannot be stacked as an RGB color image  without caution. Nevertheless, this composite image contains relevant information that can help discriminating between several classes. In particular, the DSM includes the height information which is of first importance to distinguish a roof from a road section, or a bush from a tree. Therefore, we will explore how to process these heterogeneous channels and to combine them to improve the model prediction by fusing the predictions of two networks sharing the same topology.

\subsection{Network architecture}

\subsubsection{SegNet}
\begin{figure}[t]
  \centering
  \includegraphics[width=\textwidth]{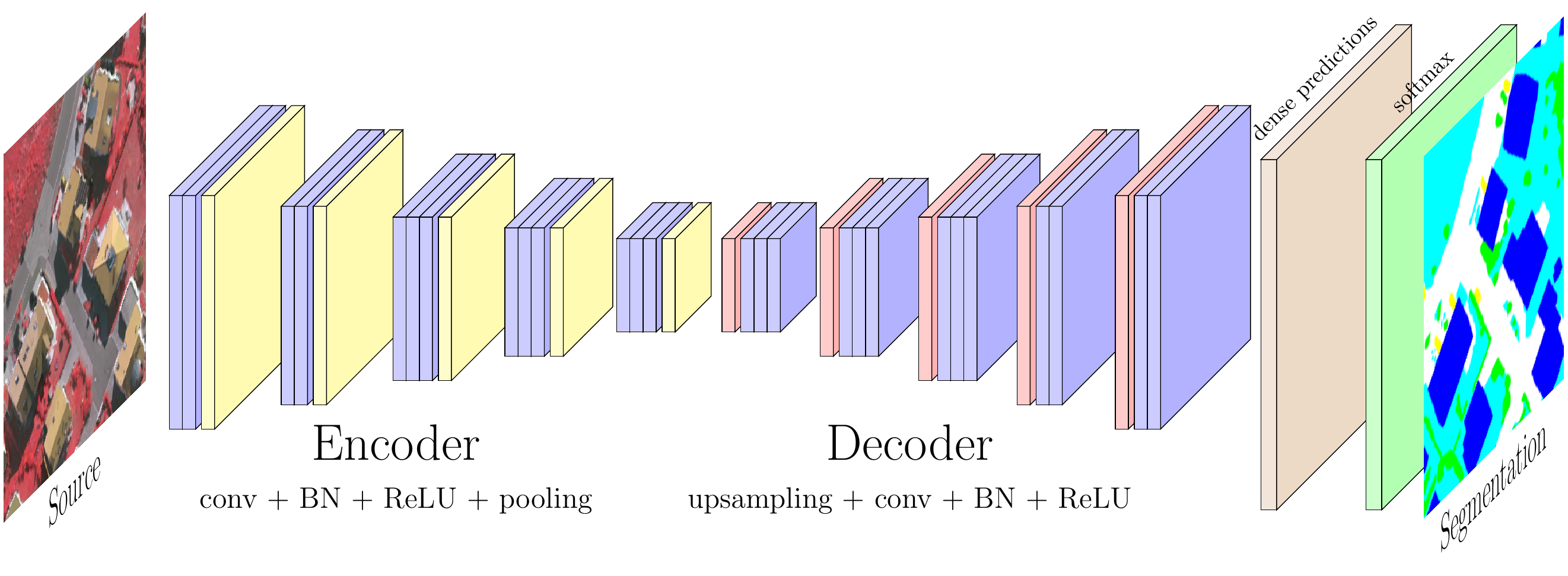}
  \caption{Illustration of the SegNet architecture applied to EO data.}
  \label{fig:segnet}
\end{figure}

There are many available architectures for semantic segmentation. We choose here the SegNet architecture~\cite{badrinarayanan_segnet:_2015} (cf. \cref{fig:segnet}), since it provides a good balance between accuracy and computational cost. SegNet's symmetrical architecture and its use of the pooling/unpooling combination is very effective for precise relocalisation of features, which is intuitively crucial for EO data. In addition to SegNet, we have performed preliminary experiments with FCN~\cite{long_fully_2015} and DeepLab~\cite{chen_semantic_2015}. Results reported no significant improvement (or even no improvement at all). Thus the need to switch to more computationally expensive architectures was not demonstrated. Note that our contributions could easily be adapted to other architectures and are not specific to SegNet.

SegNet has an encoder-decoder architecture based on the convolutional layers of VGG-16 from the Visual Geometry Group~\cite{chatfield_return_2014,simonyan_very_2014}. The encoder is a succession of convolutional layers followed by batch normalization~\cite{ioffe_batch_2015} and rectified linear units. Blocks of convolution are followed by a pooling layer of stride 2. 
The decoder has the same number of convolutions and the same number of blocks. In place of pooling, the decoder performs upsampling using unpooling layers. This layer operates by relocating at the maximum index computed by the associated pooling layer. For example, the first pooling layer computes the mask of the maximum activations (the ``$argmax$'') and passes it to the last unpooling layer, that will upsample the feature map to a full resolution by placing the activations on the mask indices and zeroes everywhere else. The sparse feature maps are then densified by the consecutive convolutional layers.
The encoding weights are initialized using the corresponding layers from VGG-16 and the decoding weights are initialized randomly using the strategy from~\cite{he_delving_2015}. We report no gain with alternative transfer functions such as ELU~\cite{clevert_fast_2015} or PReLU~\cite{he_delving_2015} and do not alter further the SegNet architecture. Let $N$ be the number of pixels in a patch and $k$ the number of classes, for a specified pixel $i$, let $y^i$ denote its label and $(z^i_1,\dots, z^i_k)$ the prediction vector; we minimize the normalized sum of the multinomial logistic loss of the softmax outputs over the whole patch:
\begin{equation}
loss = \frac{1}{N} \sum_{i=1}^N \sum_{j=1}^k y_j^i \log\left(\frac{\exp(z_j^i)}{\sum\limits_{l=1}^k \exp(z_l^i)}\right)~.
\end{equation}

As previously demonstrated in~\cite{marmanis_deep_2016}, visual filters learnt on generic datasets such as ImageNet can be effectively transferred on EO data. However, we suggest that remote sensing images have a common underlying spatial structure linked to the orthogonal line of view from the sky. Therefore, it is interesting to allow the filters to be optimized according to these specificities in order to leverage the common properties of all EO images, rather than waste parameters on useless filters. To assess this hypothesis, we experiment different learning rates for the encoder ($lr_{e}$) and the decoder ($lr_{d}$). Four strategies have been experimented:
\begin{itemize}
  \item same learning rate for both: $lr_{d} = lr_{e}$, ${lr_{e} / lr_{d}} = 1$,
  \item slightly higher learning rate for the decoder: $lr_{d} = 2 \times lr_{e}$, ${lr_{e} / lr_{d}} = 0.5$,
  \item strongly higher learning rate for the decoder: $lr_{d} = 10 \times lr_{e}$, ${lr_{e} / lr_{d}} = 0.1$,
  \item no backpropagation at all for the encoder: $lr_{e} = 0$, ${lr_{e} / lr_{d}} = 0$.
\end{itemize}
As a baseline, we also try to randomly initialize the weights of both the encoder and the decoder to train a new SegNet from scratch using the same learning rates for both parts.

\subsubsection{Multi-kernel Convolutional Layer}

\begin{figure}[t]
  \centering
  \includegraphics[width=\textwidth]{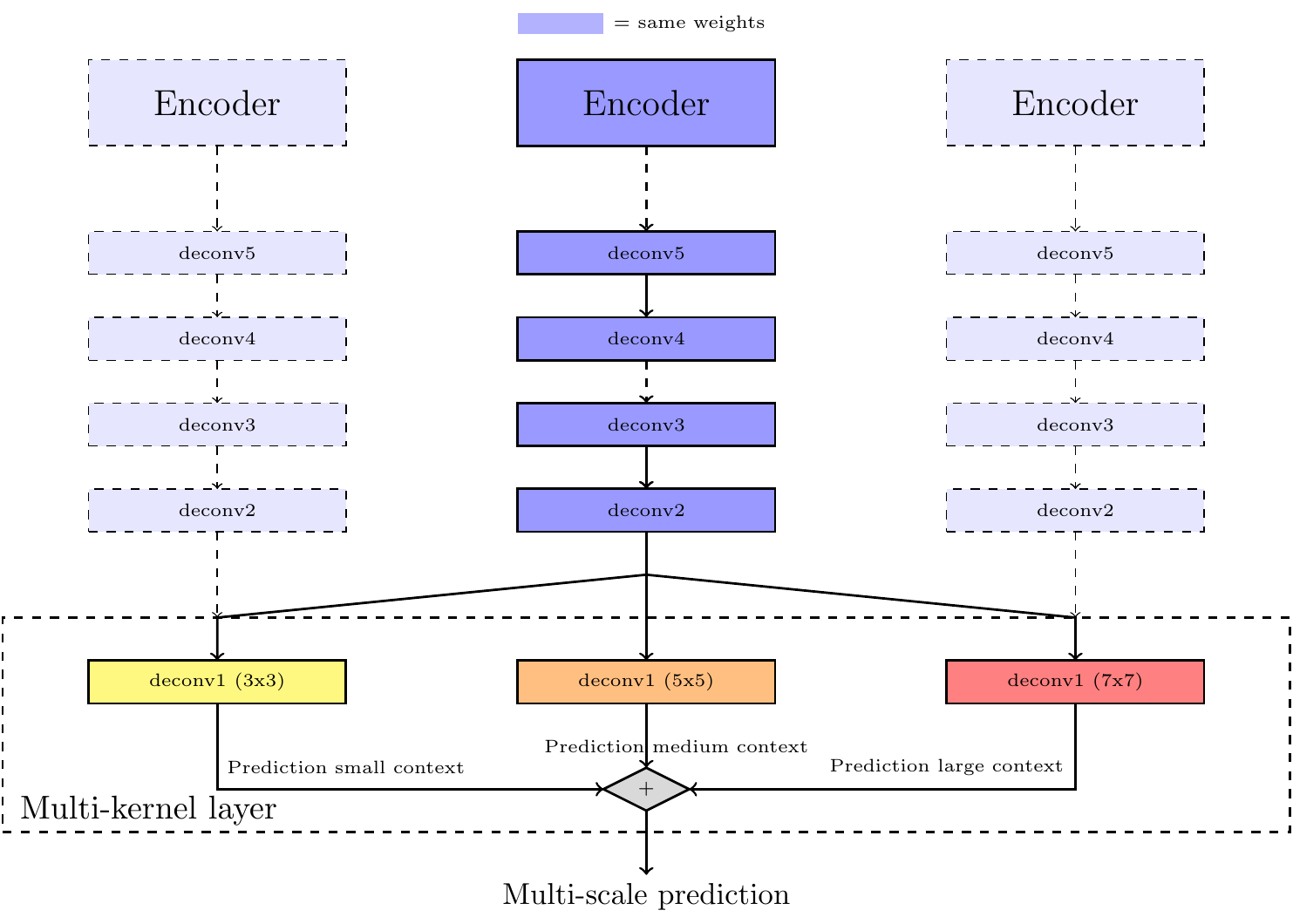}
  \caption{Our multi-kernel convolutional layer operates at 3 multiple scales, which is equivalent to averaging an ensemble of 3 models sharing weights.}
  \label{fig:contextual_module}
\end{figure}

Finally, we explore how to take spatial context into account. Let us recall that spatial information is crucial when dealing with EO data. Multi-scale processing has been proven effective for classification, notably in the Inception network~\cite{szegedy_going_2015}, for semantic segmentation~\cite{yu_multi-scale_2015} and on remote sensing imagery~\cite{zhao_learning_2016}. We design here an alternative decoder whose last layer extracts information simultaneously at several spatial resolutions and aggregates the predictions. Instead of using only one kernel size of $3\times3$, our multi-kernel convolutional layer performs 3 parallel convolutions using kernels of size $3\times3$, $5\times5$ and $7\times7$ with appropriate padding to keep the image dimensions. These different kernel sizes make possible to aggregate predictions using different receptive cell sizes. This can be seen as performing ensemble learning where the models have the same topologies and weights, excepted for the last layer, as illustrated by \cref{fig:contextual_module}. Ensemble learning with CNN has been proven to be effective in various situations, including super-resolution~\cite{liao_video_2015} where multiple CNN are used before the final deconvolution. By doing so, we are able to aggregate predictions at different scales, thus smoothing the predictions by combining different fields of view and taking into account different sizes of spatial context. If $X_p$ denotes the input activations of the multi-kernel convolutional layer for the $p^{th}$ feature map, $Z_p^s$ the activations after the convolution at the $s^{th}$ scale ($s \in \{ 1, \dots, S \}$ with $S = 3$ here), $Z'_q$ the final outputs and $W_{p,q}^s$ the $q^{th}$ convolutional kernel for the input map $p$ at scale $s$, we have:

\begin{equation}
Z'_q = \frac{1}{S} \sum_{s=1}^S Z_p^s = \frac{1}{S} \sum_{s=1}^S \sum_p W_{p,q}^s X_p~.
\end{equation}

Let $S$ denote the number of parallel convolutions (here, $S = 3$). For a given pixel at index $i$,  if $z_{k}^{s,i}$ is the activation for class $k$ and scale $s$, the logistic loss after the softmax in our multi-kernel variant is:
\begin{equation}
loss = \sum_{i=1}^N \sum_{j=1}^{k} y_j^i \log\left(\frac{\exp(\frac{1}{S} \sum\limits_{s=1}^S {z_{j}^{s,i}})}{\sum\limits_{l=1}^k \exp(\frac{1}{S} \sum\limits_{s=1}^S{z_{l}^{s,i}})}\right)~.
\end{equation}

We can train the network using the whole multi-kernel convolutional layer at once using the standard backpropagation scheme. Alternatively, we can also train only one convolution at a time, meaning that our network can be trained at first with only one scale. Then, to extend our multi-kernel layer, we can simply drop the last layer and fine-tune a new convolutional layer with another kernel size and then add the weights to a new parallel branch. This leads to a higher flexibility compared to training all scales at once, and can be used to quickly include multi-scale predictions in other fully convolutional architectures only by fine-tuning.

This multi-kernel convolutional layer shares several concepts with the competitive multi-scale convolution~\cite{liao_competitive_2015} and the Inception module~\cite{szegedy_going_2015}. However, in our work, the parallel convolutions are used only in the last layer to perform model averaging over several scales, reducing the number of parameters to be optimized compared to performing multi-scale in every layer. Moreover, this ensures more flexibility, since the number of parallel convolutions can be simply extended by fine-tuning with a new kernel size. Compared to the multi-scale context aggregation from Yu and Koltun~\cite{yu_multi-scale_2015}, our multi-kernel does not reduce dimensions and operates convolutions in parallel. Fast ensemble learning is then performed with a very low computational overhead. As opposed to Zhao et al. ~\cite{zhao_learning_2016}, we do not need to extract the patches using a pyramid, nor do we need to choose the scales beforehand, as we can extend the network according to the dataset.

\subsection{Heterogeneous Data Fusion with Residual Correction}

\begin{figure}[t]
  \centering
  \savebox{\dualsegnet}{\includegraphics[width=0.49\textwidth]{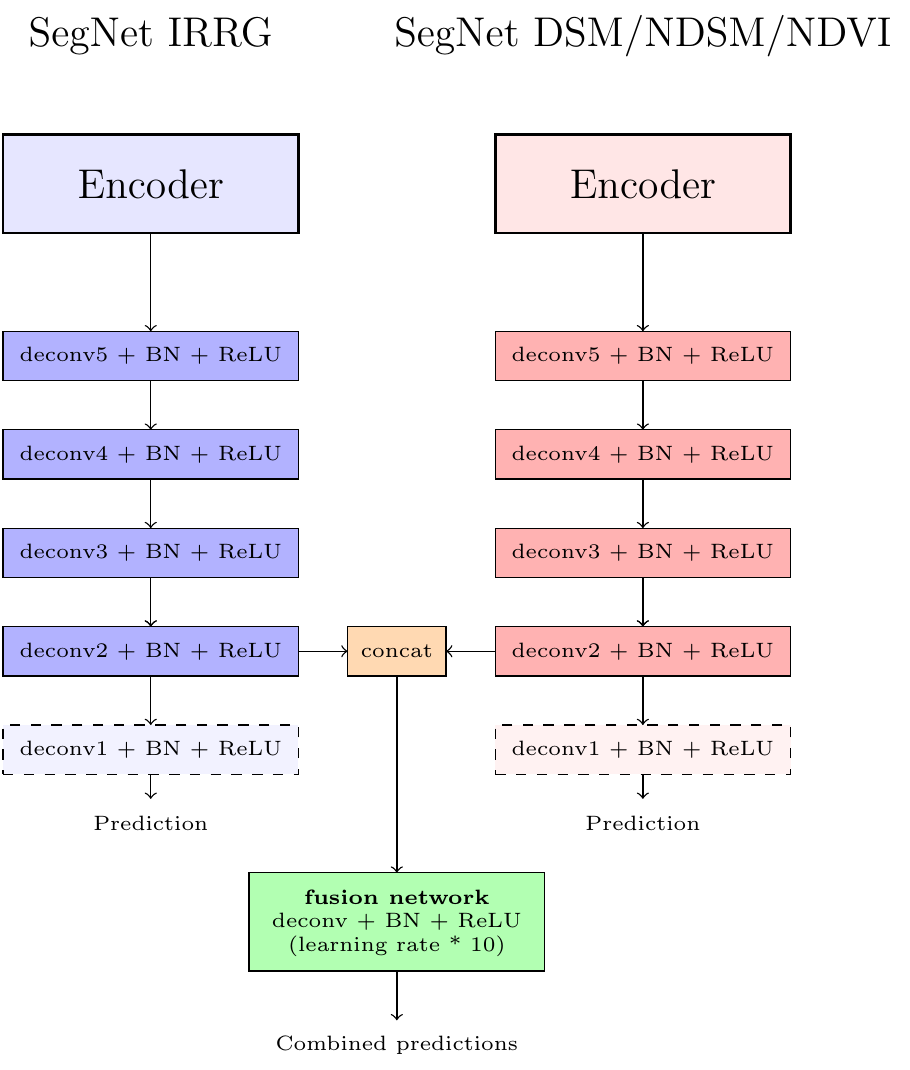}}
    \begin{subfigure}[b]{0.49\textwidth}
        \centering
        \raisebox{\dimexpr.5\ht\dualsegnet-0.42\height}{
          \includegraphics[width=\textwidth]{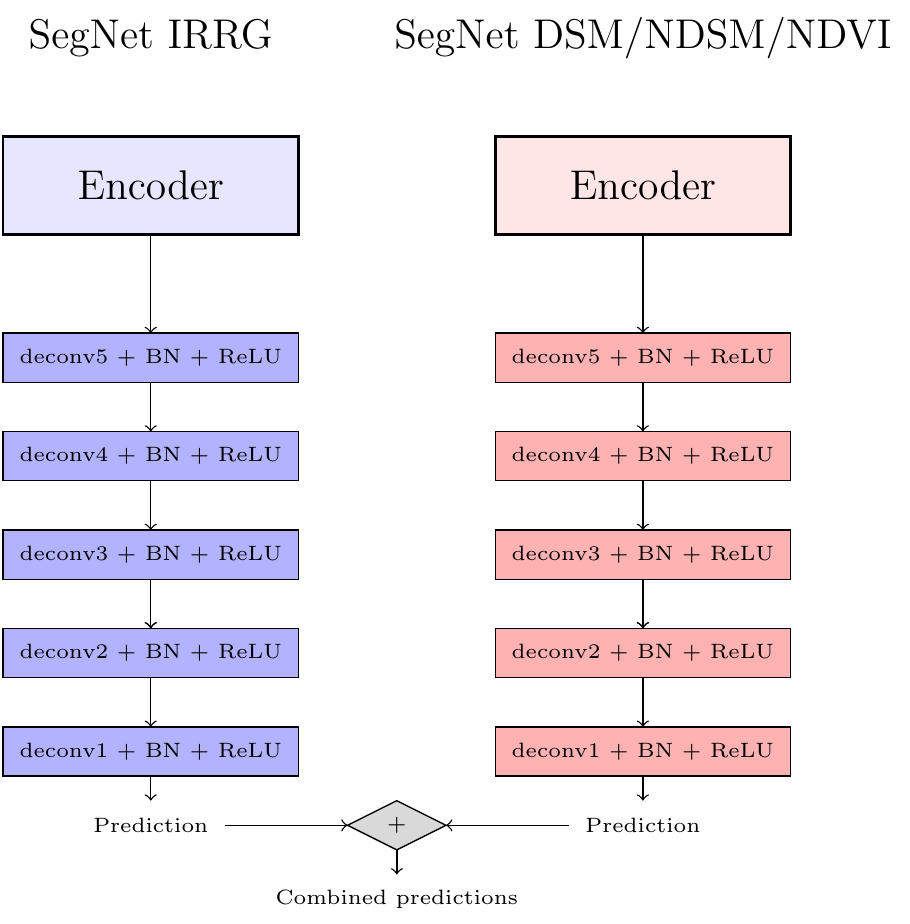}}
        \caption{Averaging strategy.}
        \label{fig:fusion_segnet_sum}
  \end{subfigure}
  \begin{subfigure}[b]{0.49\textwidth}
        \centering
        \usebox{\dualsegnet}
        \caption{Fusion network strategy.}
        \label{fig:fusion_segnet_network}
  \end{subfigure}
  \caption{Fusion strategies of our dual-stream SegNet architecture.}
  \label{fig:fusion_segnet}
\end{figure}

\begin{figure}[t]
  \centering
  \includegraphics[height=0.9\textwidth,angle=90]{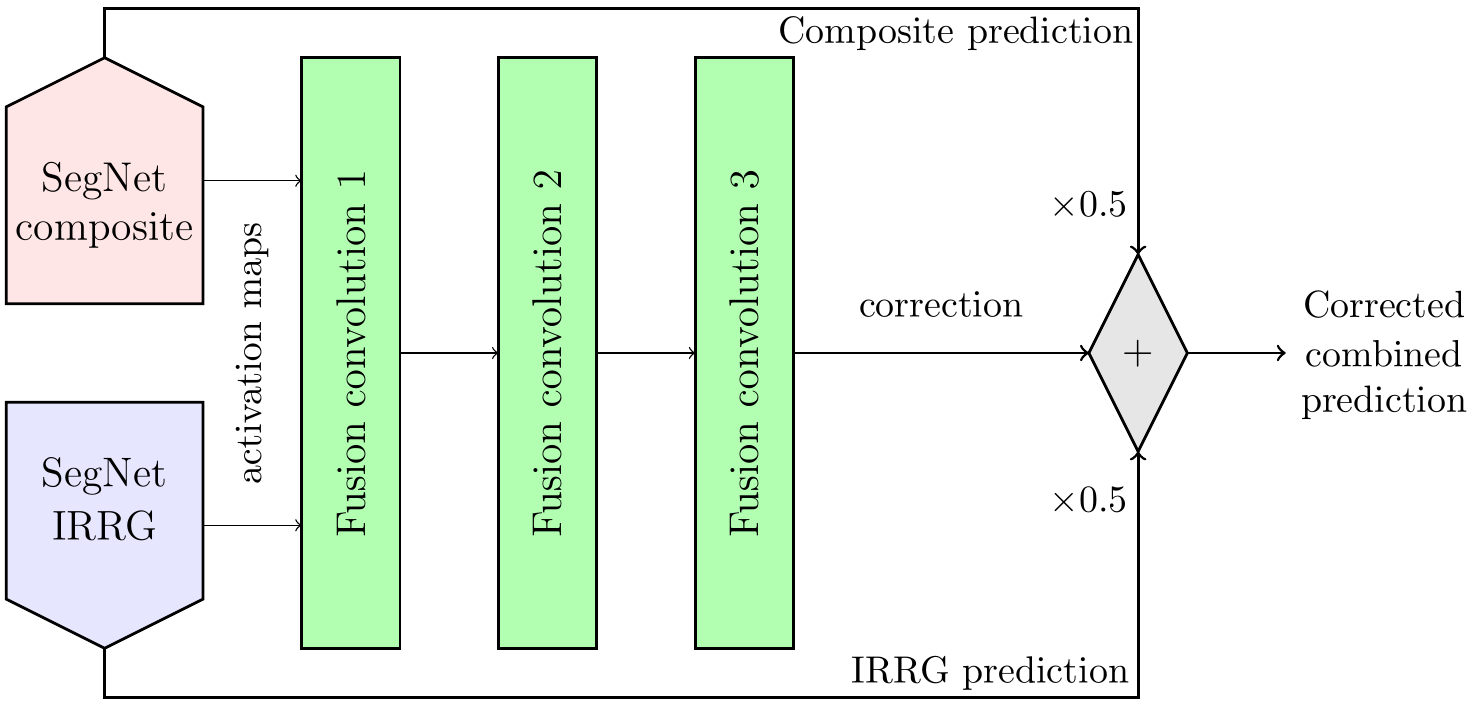}
  \caption{Fusion network for correcting the predictions using information from two complementary SegNet using heterogeneous data.}
  \label{fig:correction_network}
\end{figure}

Traditional 3-channel color images are only one possible type of remote sensing data. Multispectral sensors typically provide 4 to 12 bands, while hyperspectral images are made of a few hundreds of spectral bands. Besides, other data types such as DSM or radar imagery may be available. As stated in \cref{sec:data_proc}, IRRG data from the ISPRS dataset is completed by DSM, NDSM and NDVI. So we will assess if it is possible to: 1) build a second SegNet that can perform semantic segmentation using a second set of raw features, 2) combine the two networks to perform data fusion and improve the accuracy.

The naive data fusion would be to concatenate all 6 channels (IR/R/G and DSM/NDSM/NDVI) and feed a SegNet-like architecture with it. However, we were not able to improve the performance in regard to a simple IRRG architecture. Inspired by the multimodal fusion introduced in~\cite{ngiam_multimodal_2011} for joint audio-video representation learning and the RGB-D data fusion in~\cite{eitel_multimodal_2015}, we try a prediction-oriented fusion by merging the output activations maps. We consider here two strategies: 1) simple averaging after the softmax (\cref{fig:fusion_segnet_sum}), 2) neural network merge (\cref{fig:fusion_segnet_network}). The latter uses a corrector network that can learn from both sets of activations to correct small deficiencies in the prediction and hopefully globally improve the prediction accuracy.

Our original fusion network consisted in three convolutional layers which input was intermediate feature maps from the original network. More precisely, in the idea of fine-tuning by dropping the last fully connected layer before the softmax, we remove the last convolutional layer of each network and replace them by the fusion network convolutional layer, taking the concatenated intermediate feature maps in input. This allows the fusion network to have more information about raw activations, rather than just stacking the layers after the preprocessed predictions. Indeed, because of the one-hot encoding of the ground truth labels, the last layer activations tend to be sparse, therefore losing information about activations unrelated to the highest predicted class. However, this architecture does not improve significantly the accuracy compared to a simple averaging.

Building on the idea of residual deep learning~\cite{he_deep_2016}, we propose a fusion network based on residual correction. Instead of dropping entirely the last convolutional layers from the two SegNets, we keep them to compute the average scores. Then, we use the intermediate feature maps as inputs to a 3-convolution layers ``correction'' network, as illustrated in \cref{fig:correction_network}. Using residual learning makes sense in this case, as the average score is already a good estimation of the reality. To improve the results, we aim to use the complementary channels to correct small errors in the prediction maps. In this context, residual learning can be seen as learning a corrective term for our predictive model. Let $M_r$ denote the input of the $r^{th}$ stream ($r \in \{1,\dots,R\}$ with $R = 2$ here), $P_{r}$ the output probability tensor and $Z_r$ the intermediate feature map used for the correction. The corrected prediction is:

\begin{equation}
P'(M_1, \dots, M_R) = P(M_1, \dots, M_R) + correction(Z_1, \dots, Z_R)
\end{equation}
where
\begin{equation}
P(M_1, \dots, M_R) = \frac{1}{R}\sum_{r=1}^R P_r(M_r)~.
\end{equation}

Using residual learning should bring $\| correction \| \ll \| P \|$. This means that it should be easier for the network to learn not to add noise to predictions where its confidence is high ($ \| correction \| \simeq 0$) and only modify unsure predictions. The residual correction network can be trained by fine-tuning as usual with a logistic loss after a softmax layer.

\section{Experiments}

\subsection{Experimental Setup}

To compare our method with the current state-of-the-art, we train a model using the full dataset (training and validation sets) with the same training strategy. This is the model that we tested against other methods using the ISPRS evaluation benchmark\footnote{In this benchmark, the evaluation is not performed by us or any other competing team, but directly by the benchmark organizers.}.

\subsection{Results}

\begin{table}[t]
  \centering
  \caption{Results on the validation set with different initialization policies.}
  \begin{tabularx}{\textwidth}{ c | Y | Y | Y | Y | Y }
  \toprule
  Initialization & Random & \multicolumn{4}{c}{VGG-16}\\
  \midrule
  Learning rate ratio $\frac{lr_{e}}{lr_{d}}$ & 1 & 1 & 0.5 & 0.1 & 0\\
  \midrule
  Accuracy & 87.0\% & 87.2\% & \textbf{87.8\%} & 86.9\% & 86.5\%\\
  \bottomrule
  \end{tabularx}
  \label{tab:initialization_results}
\end{table}

\begin{table}[t]
  \centering
  \caption{Results on the validation set.}
  \setlength{\tabcolsep}{4pt}
  \begin{tabularx}{\textwidth}{ Y c c c }
  \toprule
  Type/Stride (px) & 128 {\small (no overlap)} & 64 {\small (50\% overlap)} & 32 {\small (75\% overlap)}\\
  \midrule
  Standard & 87.8\% & 88.3\% & 88.8\%\\
  Multi-kernel & 88.2\% & 88.6\% & 89.1\%\\
  Fusion (average) & 88.2\% & 88.7\% & 89.1\%\\
  Fusion (correction) & 88.6\% & 89.0\% & 89.5\%\\
  Multi-kernel + Average & 88.5\% & 89.0\% & 89.5\%\\
  Multi-kernel + Correction & 88.7\% & 89.3\% & \textbf{89.8\%}\\
  \bottomrule
  \end{tabularx}
  \label{tab:validation_results}
\end{table}

\begin{table}[t]
  \centering
  \caption{ISPRS 2D Semantic Labeling Challenge Vaihingen results.}
  \rowcolors{2}{gray!15}{white}
  \begin{tabularx}{\textwidth}{ Y c c c c c c }
  \toprule
  Method & imp surf & building & low veg & tree & car & Accuracy \\
  \midrule
  Stair Vision Library {\scriptsize (``SVL\_3'')}\cite{gerke_use_2015} & 86.6\% &	91.0\% &	77.0\% &	85.0\%	& 55.6\% &	84.8\% \\
  RF + CRF {\scriptsize (``HUST'')}\cite{quang_efficient_2015} & 86.9\% & 92.0\% &	78.3\% &	86.9\% &	29.0\% &	85.9\% \\
  CNN ensemble {\scriptsize (``ONE\_5'')}\cite{boulch_dag_2015} & 87.8\% &	92.0\% &	77.8\% &	86.2\% &	50.7\% &	85.9\% \\
  FCN {\scriptsize (``UZ\_1'')} & 89.2\% &	92.5\% &	81.6\% &	86.9\% &	57.3\% &	87.3\% \\
  FCN {\scriptsize (``UOA'')}\cite{lin_efficient_2015} & 89.8\% &	92.1\% &	80.4\% &	88.2\% &	82.0\% &	87.6\% \\
  CNN + RF + CRF {\scriptsize (``ADL\_3'')}\cite{paisitkriangkrai_effective_2015} & 89.5\% &	93.2\% &	82.3\% &	88.2\% &	63.3\% &	88.0\% \\
  FCN {\scriptsize (``DLR\_2'')}\cite{marmanis_semantic_2016} & 90.3\% &	92.3\% &	82.5\% &	89.5\% &	76.3\% &	88.5\% \\
  FCN + RF + CRF {\scriptsize (``DST\_2'')} & 90.5\% &	93.7\% &	83.4\% &	89.2\% &	72.6\% &	89.1\% \\
  \midrule
  Ours (multi-kernel) & \textbf{91.5}\% &	94.3\% &	82.7\% &	89.3\% &	\textbf{85.7}\% &	89.4\% \\
  Ours (multi-kernel + fusion) & 91.0\% &	\textbf{94.5}\% &	\textbf{84.4}\% &	\textbf{89.9}\% &	77.8\% &	\textbf{89.8}\% \\
  \bottomrule
  \end{tabularx}
  \label{tab:leaderboard}
\end{table}

\begin{figure}[t]
  \centering
  \captionsetup[subfigure]{singlelinecheck=off,justification=centering}
  \captionsetup[subfigure]{labelformat=empty}
  \begin{subfigure}[t]{0.19\textwidth}
    \includegraphics[width=\textwidth]{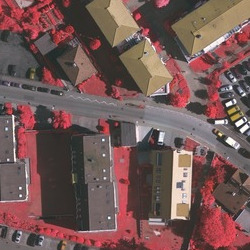}
    \caption{IRRG data}
  \end{subfigure}  \begin{subfigure}[t]{0.19\textwidth}
    \includegraphics[width=\textwidth]{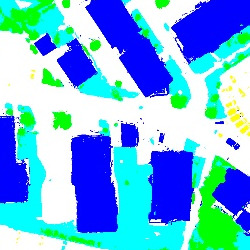}
    \caption{``SVL''\cite{gerke_use_2015}}
  \end{subfigure}
    \begin{subfigure}[t]{0.19\textwidth}
    \includegraphics[width=\textwidth]{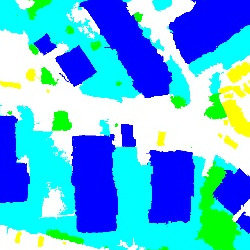}
    \caption{RF + CRF\cite{quang_efficient_2015}}
  \end{subfigure}
    \begin{subfigure}[t]{0.19\textwidth}
    \includegraphics[width=\textwidth]{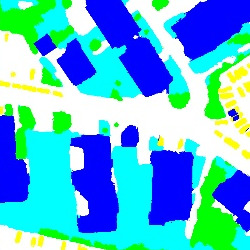}
    \caption{``DLR'' (FCN)\cite{marmanis_semantic_2016}}
  \end{subfigure}
    \begin{subfigure}[t]{0.19\textwidth}
    \includegraphics[width=\textwidth]{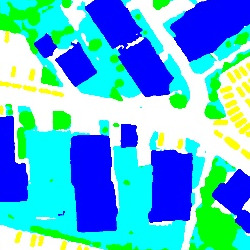}
    \caption{Ours (SegNet)}
  \end{subfigure}
  \caption{Comparison of the generated segmentations using several methods of the ISPRS Vaihingen benchmark (patch extracted from the testing set).\\
  (white: roads, {\color{blue!80!black} blue}: buildings, {\color{cyan!80!black} cyan}: low vegetation, {\color{green!80!black} green}: trees, {\color{yellow!80!black} yellow}: cars)}
  \label{fig:segnet_qualitative}
\end{figure}

Our best model achieves state-of-the art results on the ISPRS Vaihingen dataset (cf. \cref{tab:leaderboard}) \footnote{\url{http://www2.isprs.org/vaihingen-2d-semantic-labeling-contest.html}}. \cref{fig:segnet_qualitative} illustrates a qualitative comparison between SegNet using our multi-kernel convolutional layer and other baseline strategies on an extract of the Vaihingen testing set. The provided metrics are the global pixel-wise accuracy and the F1 score on each class:

\begin{equation}
F1_{i} = 2~\frac{precision_{i} \times recall_{i}}{precision_{i} + recall_{i}} \text{ and } recall_i = \frac{tp_i}{C_i},~ precision_i = \frac{tp_i}{P_i}~,
\end{equation}

where $tp_i$ the number of true positives for class $i$, $C_i$ the number of pixels belonging to class $i$, and $P_i$ the number of pixels attributed to class $i$ by the model. These metrics are computed using an alternative ground truth in which the borders have been eroded by a 3px radius circle.

Previous to our submission, the best results on the benchmark were obtained by combining FCN and hand-crafted features, whereas our method does not require any prior. The previous best method using only a FCN (``DLR\_1'') reached 88.4\%, our method improving this result by 1.4\%. Earlier methods using CNN for classification obtained 85.9\% (``ONE\_5''\cite{boulch_dag_2015}) and 86.1\% (``ADL\_1''\cite{paisitkriangkrai_effective_2015}). It should be noted that we outperform all these methods, including those that use hand-crafted features and structured models such as Conditional Random Fields, although we do not use these techniques.

\subsection{Analysis}

\subsubsection{Sliding Window Overlap}

Allowing an overlap when sliding the window across the tile slows significantly the segmentation process but improves accuracy, as shown in \cref{tab:validation_results}. Indeed, if we divide the stride by $2$, the number of patches is multiplied by $4$. However, averaging several predictions on the same region helps to correct small errors, especially around the borders of each patch, which are difficult to predict due to a lack of context. We find that a stride of 32px (75\% overlap) is fast enough for most purposes and achieves a significant boost in accuracy (+1\% compared to no overlap). Processing a tile takes 4 minutes on a Tesla K20c with a 32px stride and less than 20 seconds with a 128px stride. The inference time is doubled using the dual-stream fusion network.

\subsubsection{Transfer Learning}

As shown in \cref{tab:initialization_results}, the model achieves highest accuracy on the validation set using a low learning rate on the encoder. This supports previous evidences hinting that fine-tuning generic filters on a specialized task performs better than training new filters form scratch. However, we suggest that a too low learning rate on the original filters impede the network from reaching an optimal bank of filters if enough data is available. Indeed, in our experiments, a very low learning rate for the encoder ($0.1$) achieves a lower accuracy than a moderate drop ($0.5$). We argue that given the size and the nature (EO data) of our dataset, it is beneficial to let the filters from VGG-16 vary as this allows the network to achieve better specialization. However, a too large learning rate brings also the risk of overfitting, as showed by our experiment. Therefore, we argue that setting a lower learning rate for the encoder part of fully convolutional architectures might act as regularizer and prevent some of the overfitting that would appear otherwise. This is similar to previous results in remote sensing~\cite{nogueira_towards_2016}, but also coherent with more generic observations~\cite{yosinski_how_2014}.

\subsubsection{Multi-kernel Convolutional Layer}

\begin{figure}[t]
  \captionsetup[subfigure]{singlelinecheck=off,justification=centering}
  \captionsetup[subfigure]{labelformat=empty}
  \begin{subfigure}[t]{0.13\textwidth}
    \includegraphics[width=\textwidth]{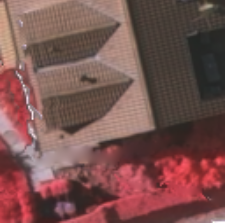}
    \caption{IRRG data}
  \end{subfigure}
  \begin{subfigure}[t]{0.13\textwidth}
    \includegraphics[width=\textwidth]{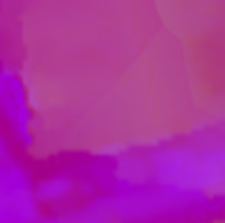}
    \caption{Composite data}
  \end{subfigure}
  \begin{subfigure}[t]{0.13\textwidth}
    \includegraphics[width=\textwidth]{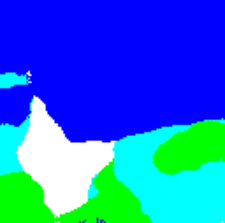}
    \caption{IRRG prediction}
  \end{subfigure}
  \begin{subfigure}[t]{0.13\textwidth}
    \includegraphics[width=\textwidth]{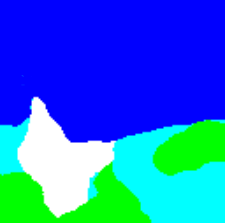}
    \caption{IRRG prediction (multi)}
  \end{subfigure}
    \begin{subfigure}[t]{0.13\textwidth}
    \includegraphics[width=\textwidth]{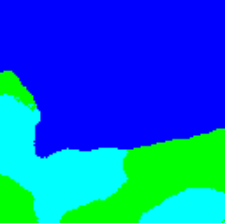}
    \caption{Composite prediction}
  \end{subfigure}
  \begin{subfigure}[t]{0.13\textwidth}
    \includegraphics[width=\textwidth]{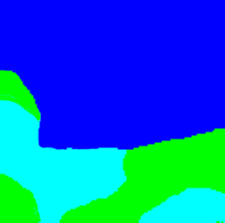}
    \caption{Composite prediction (multi)}
  \end{subfigure} 
  \begin{subfigure}[t]{0.13\textwidth}
    \includegraphics[width=\textwidth]{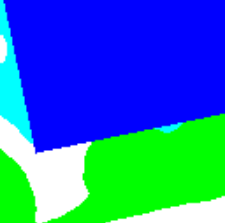}
    \caption{Ground truth}
  \end{subfigure} 
  \caption{Effects of the multi-kernel convolutional layer on selected patches.}
  \label{fig:patches_context}
\end{figure}

The multi-kernel convolutional layer brings an additional boost of 0.4\% to the accuracy. As illustrated in \cref{fig:patches_context}, it smooths the prediction by removing small artifacts isolated in large homogeneous regions. It also helps to alleviate errors by averaging predictions over several models.

This approach improves previous results on the ISPRS Vaihingen 2D labeling challenge, reaching 89.4\% \footnote{``ONE\_6'': \url{https://www.itc.nl/external/ISPRS_WGIII4/ISPRSIII_4_Test_results/2D_labeling_vaih/2D_labeling_Vaih_details_ONE_6/index.html}} (cf. \cref{tab:leaderboard}). Improvements are significant for most classes, as this multi-kernel method obtains the best F1 score for ``impervious surfaces'' (+1.0\%), ``buildings'' (+0.8\%) and ``cars'' (+3.7\%) classes. Moreover, this method is competitive on the ``low vegetation'' and ``tree'' classes. Although the cars represent only 1.2\% of the whole Vaihingen dataset and therefore does not impact strongly the global accuracy, we believe this improvement to be significant, as our model is successful both on ``stuff'' and on objects.

\subsubsection{Data Fusion and Residual Correction}

\begin{figure}[t]
  \captionsetup[subfigure]{singlelinecheck=off,justification=centering}
  \captionsetup[subfigure]{labelformat=empty}
  \centering
  \begin{subfigure}[t]{0.13\textwidth}
    \includegraphics[width=\textwidth]{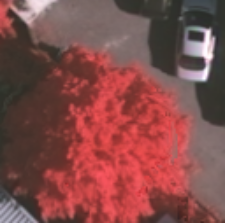}
    \caption{IRRG data}
  \end{subfigure}
  \begin{subfigure}[t]{0.13\textwidth}
    \includegraphics[width=\textwidth]{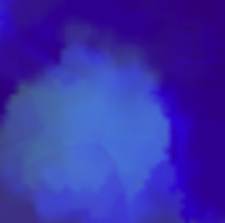}
    \caption{Composite data}
  \end{subfigure}
  \begin{subfigure}[t]{0.13\textwidth}
    \includegraphics[width=\textwidth]{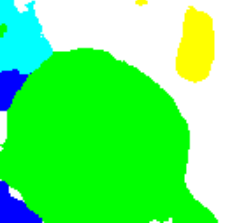}
    \caption{IRRG prediction}
  \end{subfigure}
  \begin{subfigure}[t]{0.13\textwidth}
    \includegraphics[width=\textwidth]{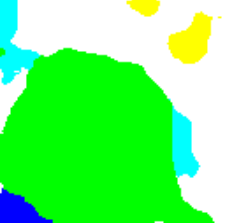}
    \caption{Composite prediction}
  \end{subfigure}
  \begin{subfigure}[t]{0.13\textwidth}
    \includegraphics[width=\textwidth]{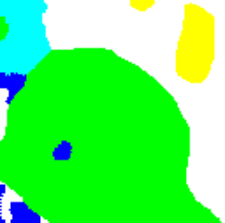}
    \caption{Fusion (average)}
  \end{subfigure}
  \begin{subfigure}[t]{0.13\textwidth}
    \includegraphics[width=\textwidth]{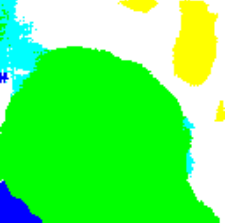}
    \caption{Fusion (network)}
  \end{subfigure} 
  \begin{subfigure}[t]{0.13\textwidth}
    \includegraphics[width=\textwidth]{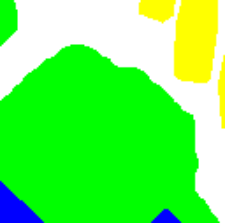}
    \caption{Ground truth}
  \end{subfigure}
    \begin{subfigure}[t]{0.13\textwidth}
    \includegraphics[width=\textwidth]{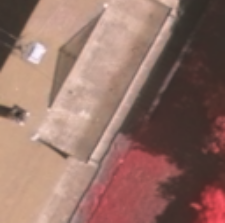}
    \caption{IRRG data}
  \end{subfigure}
  \begin{subfigure}[t]{0.13\textwidth}
    \includegraphics[width=\textwidth]{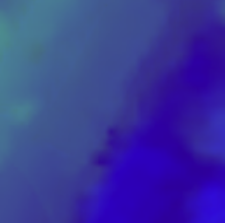}
    \caption{Composite data}
  \end{subfigure}
  \begin{subfigure}[t]{0.13\textwidth}
    \includegraphics[width=\textwidth]{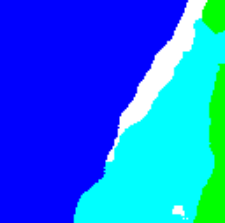}
    \caption{IRRG prediction}
  \end{subfigure}
  \begin{subfigure}[t]{0.13\textwidth}
    \includegraphics[width=\textwidth]{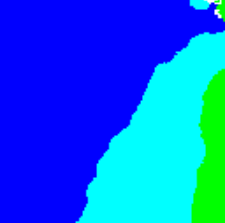}
    \caption{Composite prediction}
  \end{subfigure}
  \begin{subfigure}[t]{0.13\textwidth}
    \includegraphics[width=\textwidth]{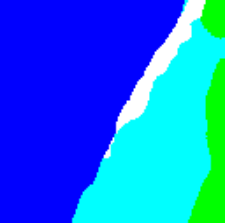}
    \caption{Fusion (average)}
  \end{subfigure}
  \begin{subfigure}[t]{0.13\textwidth}
    \includegraphics[width=\textwidth]{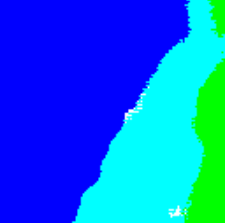}
    \caption{Fusion (network)}
  \end{subfigure} 
  \begin{subfigure}[t]{0.13\textwidth}
    \includegraphics[width=\textwidth]{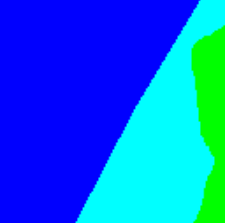}
    \caption{Ground truth}
  \end{subfigure}  
  
  \caption{Effects of our fusion strategies on selected patches.}
  \label{fig:patches_fusion}
\end{figure}

Naive prediction fusion by averaging the maps boosts the accuracy by 0.3-0.4\%. This is cumulative with the gain from the multi-kernel convolutions, which hints that the two methods are complementary. This was expected, as the latter leverages multi-scale predictions whereas the data fusion uses additional information to refine the predictions. As illustrated in \cref{fig:patches_fusion}, the fusion manages to correct errors in one model by using information from the other source. The residual correction network generates more visually appealing predictions, as it learns which network to favor for each class. For example, the IRRG data is nearly always right when predicting car pixels, therefore the correction network often keeps those. However the composite data has the advantage of the DSM to help distinguishing between low vegetation and trees. Thus, the correction network gives more weight to the predictions of the ``composite SegNet'' for these classes. Interestingly, if $m_{avg}$, $m_{corr}$, $s_{avg}$ and $s_{corr}$ denote the respective mean and standard deviation of the activations after averaging and after correction, we see that $m_{avg} \simeq 1.0,~m_{corr} \simeq 0 \text{ and } s_{avg} \simeq 5,~s_{corr} \simeq 2~$. We conclude that the network actually learnt how to apply small corrections to achieve a higher accuracy, which is in phase with both our expectations and theoretical developments~\cite{he_deep_2016}.

This approach improves our results on the ISPRS Vaihingen 2D Labeling Challenge even further, reaching 89.8\% \footnote{``ONE\_7'': \url{https://www.itc.nl/external/ISPRS_WGIII4/ISPRSIII_4_Test_results/2D_labeling_vaih/2D_labeling_Vaih_details_ONE_7/index.html}} (cf. \cref{tab:leaderboard}). F1 scores are significantly improved on buildings and vegetation, thanks to the discriminative power of the DSM and NDVI. However, even though the F1 score on cars is competitive, it is lower than expected. We explain this by the poor accuracy of the composite SegNet on cars, that degrades the average prediction and is only partly corrected by the network. We wish to investigate this issue further in the future.

\section{Conclusion and Future Work}
In this work, we investigated the use of DFCN for dense scene labeling of EO images. Especially, we showed that encoder-decoder architectures, notably SegNet, designed for semantic segmentation of traditional images and trained with weights from ImageNet, can easily be transposed to remote sensing data. This reinforces the idea that deep features and visual filters from generic images can be built upon for remote sensing tasks. We introduced in the network a multi-kernel convolutional layer that performs convolutions with several filter sizes to aggregate multi-scale predictions. This improves accuracy by performing model averaging with different sizes of spatial context. We investigated prediction-oriented data fusion with a dual-stream architecture. We showed that a residual correction network can successfully identify and correct small errors in the prediction obtained by the naive averaging of predictions coming from heterogeneous inputs. To demonstrate the relevance of those methods, we validated our methods on the ISPRS 2D Vaihingen semantic labeling challenge, on which we improved the state-of-the-art by~1\%.

In the future, we would like to investigate if residual correction can improve performance for networks with different topologies. Moreover, we hope to study how to perform data-oriented fusion, sooner in the network, to reduce the computational overhead of using several long parallel streams. Finally, we believe that there is additional progress to be made by integrating the multi-scale nature of the data early in the network design.

\vspace{3mm}
\noindent {\bf Acknowledgement}. The Vaihingen data set was provided by the German Society for Photogrammetry, Remote Sensing and Geoinformation (DGPF) \cite{cramer_dgpf_2010}: \url{http://www.ifp.uni-stuttgart.de/dgpf/DKEP-Allg.html}.

Nicolas Audebert's work is supported by the Total-ONERA research project NAOMI. The authors acknowledge the support of the French Agence Nationale de la Recherche (ANR) under reference ANR-13-JS02-0005-01 (Asterix project).

\bibliographystyle{splncs}
\bibliography{ACCV16}

\providecommand*{\mcitethebibliography}{\thebibliography}
\csname @ifundefined\endcsname{endmcitethebibliography}
{\let\endmcitethebibliography\endthebibliography}{}
\begin{mcitethebibliography}{54}
\providecommand*{\natexlab}[1]{#1}
\providecommand*{\mciteSetBstSublistMode}[1]{}
\providecommand*{\mciteSetBstMaxWidthForm}[2]{}
\providecommand*{\mciteBstWouldAddEndPuncttrue}
  {\def\EndOfBibitem{\unskip.}}
\providecommand*{\mciteBstWouldAddEndPunctfalse}
  {\let\EndOfBibitem\relax}
\providecommand*{\mciteSetBstMidEndSepPunct}[3]{}
\providecommand*{\mciteSetBstSublistLabelBeginEnd}[3]{}
\providecommand*{\EndOfBibitem}{}
\mciteSetBstSublistMode{f}
\mciteSetBstMaxWidthForm{subitem}
{(\emph{\alph{mcitesubitemcount}})}
\mciteSetBstSublistLabelBeginEnd{\mcitemaxwidthsubitemform\space}
{\relax}{\relax}

\bibitem[Lodish \emph{et~al.}(2012)Lodish\emph{et~al.}]{lodish_book}
H.~Lodish \emph{et~al.}, \emph{Molecular Cell Biology, 7th ed.}, W. H. Freeman
  \& Co, New York, 2012\relax
\mciteBstWouldAddEndPuncttrue
\mciteSetBstMidEndSepPunct{\mcitedefaultmidpunct}
{\mcitedefaultendpunct}{\mcitedefaultseppunct}\relax
\EndOfBibitem
\bibitem[Evans and Yeung(1994)]{Evans94CPL}
E.~Evans and A.~Yeung, \emph{Chem. Phys. Lipids}, 1994, \textbf{73}, 39\relax
\mciteBstWouldAddEndPuncttrue
\mciteSetBstMidEndSepPunct{\mcitedefaultmidpunct}
{\mcitedefaultendpunct}{\mcitedefaultseppunct}\relax
\EndOfBibitem
\bibitem[Seifert and Langer(1993)]{Seifert93EPL}
U.~Seifert and S.~A. Langer, \emph{Europhys. Lett.}, 1993, \textbf{23},
  71\relax
\mciteBstWouldAddEndPuncttrue
\mciteSetBstMidEndSepPunct{\mcitedefaultmidpunct}
{\mcitedefaultendpunct}{\mcitedefaultseppunct}\relax
\EndOfBibitem
\bibitem[Fournier \emph{et~al.}(2009)Fournier, Khalifat, Puff, and
  Angelova]{Fournier09PRL}
J.-B. Fournier, N.~Khalifat, N.~Puff and M.~I. Angelova, \emph{Phys. Rev.
  Lett.}, 2009, \textbf{102}, 018102\relax
\mciteBstWouldAddEndPuncttrue
\mciteSetBstMidEndSepPunct{\mcitedefaultmidpunct}
{\mcitedefaultendpunct}{\mcitedefaultseppunct}\relax
\EndOfBibitem
\bibitem[Brennen(2005)]{Brennen_book}
C.~E. Brennen, \emph{Fundamentals of Multiphase Flow}, Cambridge University
  Press, New York, 2005\relax
\mciteBstWouldAddEndPuncttrue
\mciteSetBstMidEndSepPunct{\mcitedefaultmidpunct}
{\mcitedefaultendpunct}{\mcitedefaultseppunct}\relax
\EndOfBibitem
\bibitem[Saffman and Delbr{\"u}ck(1975)]{Saffman:1975aa}
P.~G. Saffman and M.~Delbr{\"u}ck, \emph{Proc Natl Acad Sci U S A}, 1975,
  \textbf{72}, 3111--3\relax
\mciteBstWouldAddEndPuncttrue
\mciteSetBstMidEndSepPunct{\mcitedefaultmidpunct}
{\mcitedefaultendpunct}{\mcitedefaultseppunct}\relax
\EndOfBibitem
\bibitem[Reister and Seifert(2005)]{Reister:2005}
E.~Reister and U.~Seifert, \emph{EPL (Europhysics Letters)}, 2005, \textbf{71},
  859\relax
\mciteBstWouldAddEndPuncttrue
\mciteSetBstMidEndSepPunct{\mcitedefaultmidpunct}
{\mcitedefaultendpunct}{\mcitedefaultseppunct}\relax
\EndOfBibitem
\bibitem[Leitenberger \emph{et~al.}(2008)Leitenberger, Reister-Gottfried, and
  Seifert]{Leitenberger:2008aa}
S.~M. Leitenberger, E.~Reister-Gottfried and U.~Seifert, \emph{Langmuir}, 2008,
  \textbf{24}, 1254--61\relax
\mciteBstWouldAddEndPuncttrue
\mciteSetBstMidEndSepPunct{\mcitedefaultmidpunct}
{\mcitedefaultendpunct}{\mcitedefaultseppunct}\relax
\EndOfBibitem
\bibitem[Naji \emph{et~al.}(2009)Naji, Atzberger, and Brown]{Naji:2009aa}
A.~Naji, P.~J. Atzberger and F.~L.~H. Brown, \emph{Phys Rev Lett}, 2009,
  \textbf{102}, 138102\relax
\mciteBstWouldAddEndPuncttrue
\mciteSetBstMidEndSepPunct{\mcitedefaultmidpunct}
{\mcitedefaultendpunct}{\mcitedefaultseppunct}\relax
\EndOfBibitem
\bibitem[Reister-Gottfried \emph{et~al.}(2010)Reister-Gottfried, Leitenberger,
  and Seifert]{Reister-Gottfried:2010aa}
E.~Reister-Gottfried, S.~M. Leitenberger and U.~Seifert, \emph{Phys Rev E Stat
  Nonlin Soft Matter Phys}, 2010, \textbf{81}, 031903\relax
\mciteBstWouldAddEndPuncttrue
\mciteSetBstMidEndSepPunct{\mcitedefaultmidpunct}
{\mcitedefaultendpunct}{\mcitedefaultseppunct}\relax
\EndOfBibitem
\bibitem[Quemeneur \emph{et~al.}(2014)Quemeneur\emph{et~al.}]{Quemeneur14}
F.~Quemeneur \emph{et~al.}, \emph{PNAS}, 2014, \textbf{111}, 5083\relax
\mciteBstWouldAddEndPuncttrue
\mciteSetBstMidEndSepPunct{\mcitedefaultmidpunct}
{\mcitedefaultendpunct}{\mcitedefaultseppunct}\relax
\EndOfBibitem
\bibitem[Leibler(1986)]{Leibler:1986}
S.~Leibler, \emph{Journal de Physique}, 1986, \textbf{47}, 507--516\relax
\mciteBstWouldAddEndPuncttrue
\mciteSetBstMidEndSepPunct{\mcitedefaultmidpunct}
{\mcitedefaultendpunct}{\mcitedefaultseppunct}\relax
\EndOfBibitem
\bibitem[Leibler and Andelman(1987)]{Leibler:1987}
S.~Leibler and D.~Andelman, \emph{Journal de Physique}, 1987, \textbf{48},
  2013\relax
\mciteBstWouldAddEndPuncttrue
\mciteSetBstMidEndSepPunct{\mcitedefaultmidpunct}
{\mcitedefaultendpunct}{\mcitedefaultseppunct}\relax
\EndOfBibitem
\bibitem[Camley and Brown(2013)]{Camley:2013}
B.~A. Camley and F.~L.~H. Brown, \emph{Soft Matter}, 2013, \textbf{9},
  4767--4779\relax
\mciteBstWouldAddEndPuncttrue
\mciteSetBstMidEndSepPunct{\mcitedefaultmidpunct}
{\mcitedefaultendpunct}{\mcitedefaultseppunct}\relax
\EndOfBibitem
\bibitem[Lomholt \emph{et~al.}(2005)Lomholt, Hansen, and Miao]{Lomholt:2005aa}
M.~A. Lomholt, P.~L. Hansen and L.~Miao, \emph{Eur. Phys. J. E}, 2005,
  \textbf{16}, 439--61\relax
\mciteBstWouldAddEndPuncttrue
\mciteSetBstMidEndSepPunct{\mcitedefaultmidpunct}
{\mcitedefaultendpunct}{\mcitedefaultseppunct}\relax
\EndOfBibitem
\bibitem[Kramer(1971)]{Kramer71JCP}
L.~Kramer, \emph{J. Chem. Phys.}, 1971, \textbf{55}, 2097\relax
\mciteBstWouldAddEndPuncttrue
\mciteSetBstMidEndSepPunct{\mcitedefaultmidpunct}
{\mcitedefaultendpunct}{\mcitedefaultseppunct}\relax
\EndOfBibitem
\bibitem[Brochard and Lennon(1975)]{Brochard75JPhys}
F.~Brochard and J.-F. Lennon, \emph{J. Phys.}, 1975, \textbf{36}, 1035\relax
\mciteBstWouldAddEndPuncttrue
\mciteSetBstMidEndSepPunct{\mcitedefaultmidpunct}
{\mcitedefaultendpunct}{\mcitedefaultseppunct}\relax
\EndOfBibitem
\bibitem[Helfrich(1973)]{Helfrich73}
W.~Helfrich, \emph{Z. NaturForsch.}, 1973, \textbf{C 28}, 693\relax
\mciteBstWouldAddEndPuncttrue
\mciteSetBstMidEndSepPunct{\mcitedefaultmidpunct}
{\mcitedefaultendpunct}{\mcitedefaultseppunct}\relax
\EndOfBibitem
\bibitem[Arroyo and DeSimone(2009)]{Arroyo09PRE}
M.~Arroyo and A.~DeSimone, \emph{Phys. Rev. E}, 2009, \textbf{79}, 031915\relax
\mciteBstWouldAddEndPuncttrue
\mciteSetBstMidEndSepPunct{\mcitedefaultmidpunct}
{\mcitedefaultendpunct}{\mcitedefaultseppunct}\relax
\EndOfBibitem
\bibitem[Rahimi and Arroyo(2012)]{Rahimi12PRE}
M.~Rahimi and M.~Arroyo, \emph{Phys. Rev. E}, 2012, \textbf{86}, 011932\relax
\mciteBstWouldAddEndPuncttrue
\mciteSetBstMidEndSepPunct{\mcitedefaultmidpunct}
{\mcitedefaultendpunct}{\mcitedefaultseppunct}\relax
\EndOfBibitem
\bibitem[Rahimi \emph{et~al.}(2013)Rahimi, DeSimone, and Arroyo]{Rahimi13SM}
M.~Rahimi, A.~DeSimone and M.~Arroyo, \emph{Soft Matter}, 2013, \textbf{9},
  11033\relax
\mciteBstWouldAddEndPuncttrue
\mciteSetBstMidEndSepPunct{\mcitedefaultmidpunct}
{\mcitedefaultendpunct}{\mcitedefaultseppunct}\relax
\EndOfBibitem
\bibitem[Goldstein \emph{et~al.}(2000)Goldstein, Poole, and
  Safko]{Goldstein_book}
H.~Goldstein, C.~Poole and J.~Safko, \emph{Classical Mechanics}, Addison
  Welsley, New York, 2000\relax
\mciteBstWouldAddEndPuncttrue
\mciteSetBstMidEndSepPunct{\mcitedefaultmidpunct}
{\mcitedefaultendpunct}{\mcitedefaultseppunct}\relax
\EndOfBibitem
\bibitem[Doi(2011)]{Doi_Onsager}
M.~Doi, \emph{J. Phys.: Condens. Matter}, 2011, \textbf{23}, 284118\relax
\mciteBstWouldAddEndPuncttrue
\mciteSetBstMidEndSepPunct{\mcitedefaultmidpunct}
{\mcitedefaultendpunct}{\mcitedefaultseppunct}\relax
\EndOfBibitem
\bibitem[Fournier(2015)]{Fournier15IJNM}
J.-B. Fournier, \emph{Int. J. Nonlinear Mech.}, 2015, \textbf{75}, 67\relax
\mciteBstWouldAddEndPuncttrue
\mciteSetBstMidEndSepPunct{\mcitedefaultmidpunct}
{\mcitedefaultendpunct}{\mcitedefaultseppunct}\relax
\EndOfBibitem
\bibitem[Safran(2003)]{Statistical2003Safran}
S.~A. Safran, \emph{{Statistical thermodynamics of surfaces, interfaces, and
  membranes}}, Westview Press, 2003\relax
\mciteBstWouldAddEndPuncttrue
\mciteSetBstMidEndSepPunct{\mcitedefaultmidpunct}
{\mcitedefaultendpunct}{\mcitedefaultseppunct}\relax
\EndOfBibitem
\bibitem[Svetina \emph{et~al.}(1985)Svetina, Brumen, and {\v Z}ek{\v
  s}]{Svetina85}
S.~Svetina, M.~Brumen and B.~{\v Z}ek{\v s}, \emph{Stud. Biophys.}, 1985,
  \textbf{110}, 77\relax
\mciteBstWouldAddEndPuncttrue
\mciteSetBstMidEndSepPunct{\mcitedefaultmidpunct}
{\mcitedefaultendpunct}{\mcitedefaultseppunct}\relax
\EndOfBibitem
\bibitem[Miao \emph{et~al.}(1994)Miao, Seifert, Wortis, and
  D{\"o}bereiner]{Miao94PRE}
L.~Miao, U.~Seifert, M.~Wortis and H.-G. D{\"o}bereiner, \emph{Phys. Rev. E},
  1994, \textbf{49}, 5389\relax
\mciteBstWouldAddEndPuncttrue
\mciteSetBstMidEndSepPunct{\mcitedefaultmidpunct}
{\mcitedefaultendpunct}{\mcitedefaultseppunct}\relax
\EndOfBibitem
\bibitem[Sens(2004)]{Sens04PRL}
P.~Sens, \emph{Phys. Rev. Lett.}, 2004, \textbf{93}, 108103\relax
\mciteBstWouldAddEndPuncttrue
\mciteSetBstMidEndSepPunct{\mcitedefaultmidpunct}
{\mcitedefaultendpunct}{\mcitedefaultseppunct}\relax
\EndOfBibitem
\bibitem[Bitbol \emph{et~al.}(2012)Bitbol\emph{et~al.}]{Bitbol12SM}
A.-F. Bitbol \emph{et~al.}, \emph{Soft Matter}, 2012, \textbf{8}, 6073\relax
\mciteBstWouldAddEndPuncttrue
\mciteSetBstMidEndSepPunct{\mcitedefaultmidpunct}
{\mcitedefaultendpunct}{\mcitedefaultseppunct}\relax
\EndOfBibitem
\bibitem[den Otter and Shkulipa(2007)]{Otter:2007aa}
W.~K. den Otter and S.~A. Shkulipa, \emph{Biophys J}, 2007, \textbf{93},
  423--33\relax
\mciteBstWouldAddEndPuncttrue
\mciteSetBstMidEndSepPunct{\mcitedefaultmidpunct}
{\mcitedefaultendpunct}{\mcitedefaultseppunct}\relax
\EndOfBibitem
\bibitem[Okamoto \emph{et~al.}(2015)Okamoto, Kanemori, Komura, and
  Fournier]{Okamoto15condmat}
R.~Okamoto, Y.~Kanemori, S.~Komura and J.-B. Fournier, \emph{arXiv:1511.01324
  [cond-mat.soft]}, 2015\relax
\mciteBstWouldAddEndPuncttrue
\mciteSetBstMidEndSepPunct{\mcitedefaultmidpunct}
{\mcitedefaultendpunct}{\mcitedefaultseppunct}\relax
\EndOfBibitem
\bibitem[jus()]{justif1}
The term $\sim\!\nabla^2h$ is discarded as a boundary term. The terms
  $\sim\!\rho^\pm$ and $\sim\!\phi^\pm$ are discarded as both $\int
  d^2x_\perp\, \rho^\pm$ and $\int d^2x_\perp\, (1+\rho^\pm)\phi^\pm$ are
  constant due to the mass conservation of lipids and proteins.\relax
\mciteBstWouldAddEndPunctfalse
\mciteSetBstMidEndSepPunct{\mcitedefaultmidpunct}
{}{\mcitedefaultseppunct}\relax
\EndOfBibitem
\bibitem[Brannigan and Brown(2007)]{Brannigan:2007aa}
G.~Brannigan and F.~L.~H. Brown, \emph{Biophys J}, 2007, \textbf{92},
  864--76\relax
\mciteBstWouldAddEndPuncttrue
\mciteSetBstMidEndSepPunct{\mcitedefaultmidpunct}
{\mcitedefaultendpunct}{\mcitedefaultseppunct}\relax
\EndOfBibitem
\bibitem[West \emph{et~al.}(2009)West, Brown, and Schmid]{West2009101}
B.~West, F.~L. Brown and F.~Schmid, \emph{Biophysical Journal}, 2009,
  \textbf{96}, 101 -- 115\relax
\mciteBstWouldAddEndPuncttrue
\mciteSetBstMidEndSepPunct{\mcitedefaultmidpunct}
{\mcitedefaultendpunct}{\mcitedefaultseppunct}\relax
\EndOfBibitem
\bibitem[Bitbol \emph{et~al.}(2012)Bitbol, Constantin, and
  Fournier]{Bitbol2012}
A.~F. Bitbol, D.~Constantin and J.~B. Fournier, \emph{PLoS One}, 2012,
  \textbf{7}, e48306\relax
\mciteBstWouldAddEndPuncttrue
\mciteSetBstMidEndSepPunct{\mcitedefaultmidpunct}
{\mcitedefaultendpunct}{\mcitedefaultseppunct}\relax
\EndOfBibitem
\bibitem[Watson \emph{et~al.}(2013)Watson, Morriss-Andrews, Welch, and
  Brown]{Watson2013}
M.~C. Watson, A.~Morriss-Andrews, P.~M. Welch and F.~L.~H. Brown, \emph{The
  Journal of Chemical Physics}, 2013, \textbf{139}, 084706\relax
\mciteBstWouldAddEndPuncttrue
\mciteSetBstMidEndSepPunct{\mcitedefaultmidpunct}
{\mcitedefaultendpunct}{\mcitedefaultseppunct}\relax
\EndOfBibitem
\bibitem[not()]{note_moderate}
There are lipids in between the rigid proteins, therefore bending involves
  essentially splaying the lipids and stretching involves essentially changing
  the lipid density. The presence of the proteins should thus not change
  drastically the associated energy penalty per unit area.\relax
\mciteBstWouldAddEndPunctfalse
\mciteSetBstMidEndSepPunct{\mcitedefaultmidpunct}
{}{\mcitedefaultseppunct}\relax
\EndOfBibitem
\bibitem[Landau and Lifshitz(2000)]{Landau_Fluid_Mechanics_book}
L.~D. Landau and E.~M. Lifshitz, \emph{Fluid Mechanics}, Butterworth-Heinemann,
  Oxford, 2000\relax
\mciteBstWouldAddEndPuncttrue
\mciteSetBstMidEndSepPunct{\mcitedefaultmidpunct}
{\mcitedefaultendpunct}{\mcitedefaultseppunct}\relax
\EndOfBibitem
\bibitem[Niemel\"a \emph{et~al.}(2010)Niemel\"a, Miettinen, Monticelli,
  Hammaren, Bjelkmar, Murtola, Lindahl, and Vattulainen]{Niemela}
P.~S. Niemel\"a, M.~S. Miettinen, L.~Monticelli, H.~Hammaren, P.~Bjelkmar,
  T.~Murtola, E.~Lindahl and I.~Vattulainen, \emph{Journal of the American
  Chemical Society}, 2010, \textbf{132}, 7574--7575\relax
\mciteBstWouldAddEndPuncttrue
\mciteSetBstMidEndSepPunct{\mcitedefaultmidpunct}
{\mcitedefaultendpunct}{\mcitedefaultseppunct}\relax
\EndOfBibitem
\bibitem[Oppenheimer and Diamant(2009)]{Oppenheimer:2009aa}
N.~Oppenheimer and H.~Diamant, \emph{Biophys J}, 2009, \textbf{96},
  3041--9\relax
\mciteBstWouldAddEndPuncttrue
\mciteSetBstMidEndSepPunct{\mcitedefaultmidpunct}
{\mcitedefaultendpunct}{\mcitedefaultseppunct}\relax
\EndOfBibitem
\bibitem[Camley and Brown(2014)]{Camley2014}
B.~A. Camley and F.~L.~H. Brown, \emph{The Journal of Chemical Physics}, 2014,
  \textbf{141}, 075103\relax
\mciteBstWouldAddEndPuncttrue
\mciteSetBstMidEndSepPunct{\mcitedefaultmidpunct}
{\mcitedefaultendpunct}{\mcitedefaultseppunct}\relax
\EndOfBibitem
\bibitem[Khoshnood \emph{et~al.}(2010)Khoshnood, Noguchi, and
  Gompper]{Khoshnood:2010aa}
A.~Khoshnood, H.~Noguchi and G.~Gompper, \emph{J Chem Phys}, 2010,
  \textbf{132}, 025101\relax
\mciteBstWouldAddEndPuncttrue
\mciteSetBstMidEndSepPunct{\mcitedefaultmidpunct}
{\mcitedefaultendpunct}{\mcitedefaultseppunct}\relax
\EndOfBibitem
\bibitem[Rawicz \emph{et~al.}(2000)Rawicz\emph{et~al.}]{Rawicz00BiophysJ}
W.~Rawicz \emph{et~al.}, \emph{Biophys. J.}, 2000, \textbf{79}, 328\relax
\mciteBstWouldAddEndPuncttrue
\mciteSetBstMidEndSepPunct{\mcitedefaultmidpunct}
{\mcitedefaultendpunct}{\mcitedefaultseppunct}\relax
\EndOfBibitem
\bibitem[Honerkamp-Smith \emph{et~al.}(2013)Honerkamp-Smith, Woodhouse,
  Kantsler, and Goldstein]{Honerkamp-Smith:2013aa}
A.~R. Honerkamp-Smith, F.~G. Woodhouse, V.~Kantsler and R.~E. Goldstein,
  \emph{Phys Rev Lett}, 2013, \textbf{111}, 038103\relax
\mciteBstWouldAddEndPuncttrue
\mciteSetBstMidEndSepPunct{\mcitedefaultmidpunct}
{\mcitedefaultendpunct}{\mcitedefaultseppunct}\relax
\EndOfBibitem
\bibitem[Merkel \emph{et~al.}(1989)Merkel, Sackmann, and Evans]{Merkel89JPhys}
R.~Merkel, E.~Sackmann and E.~Evans, \emph{J. Phys. France}, 1989, \textbf{50},
  1535\relax
\mciteBstWouldAddEndPuncttrue
\mciteSetBstMidEndSepPunct{\mcitedefaultmidpunct}
{\mcitedefaultendpunct}{\mcitedefaultseppunct}\relax
\EndOfBibitem
\bibitem[Pott and M\'el\'eard(2002)]{Pott02Europhys}
T.~Pott and P.~M\'el\'eard, \emph{Europhys. Lett.}, 2002, \textbf{59}, 87\relax
\mciteBstWouldAddEndPuncttrue
\mciteSetBstMidEndSepPunct{\mcitedefaultmidpunct}
{\mcitedefaultendpunct}{\mcitedefaultseppunct}\relax
\EndOfBibitem
\bibitem[Rodr{\'\i}guez-Garc{\'\i}a
  \emph{et~al.}(2009)Rodr{\'\i}guez-Garc{\'\i}a, Arriaga, Mell, Moleiro,
  L{\'o}pez-Montero, and Monroy]{Rodriguez-Garcia:2009aa}
R.~Rodr{\'\i}guez-Garc{\'\i}a, L.~R. Arriaga, M.~Mell, L.~H. Moleiro,
  I.~L{\'o}pez-Montero and F.~Monroy, \emph{Phys Rev Lett}, 2009, \textbf{102},
  128101\relax
\mciteBstWouldAddEndPuncttrue
\mciteSetBstMidEndSepPunct{\mcitedefaultmidpunct}
{\mcitedefaultendpunct}{\mcitedefaultseppunct}\relax
\EndOfBibitem
\bibitem[Ramadurai \emph{et~al.}(2009)Ramadurai\emph{et~al.}]{Ramadurai09JACS}
S.~Ramadurai \emph{et~al.}, \emph{J. Am. Chem. Soc.}, 2009, \textbf{131},
  12650\relax
\mciteBstWouldAddEndPuncttrue
\mciteSetBstMidEndSepPunct{\mcitedefaultmidpunct}
{\mcitedefaultendpunct}{\mcitedefaultseppunct}\relax
\EndOfBibitem
\bibitem[not()]{notealpha}
The mixing entropy for proteins much larger than lipids of size $a$ yields
  $\alpha=k_\mathrm{B}T/[ka^2(1-c_0)]\approx1$, for $c_0\simeq0.3$.\relax
\mciteBstWouldAddEndPunctfalse
\mciteSetBstMidEndSepPunct{\mcitedefaultmidpunct}
{}{\mcitedefaultseppunct}\relax
\EndOfBibitem
\bibitem[com()]{compareSeifert}
For $c_0=0$ we recover the dynamical matrix of Ref.~\cite{Seifert93EPL}, except
  that $\eta_s$ includes dilational viscosity.\relax
\mciteBstWouldAddEndPunctfalse
\mciteSetBstMidEndSepPunct{\mcitedefaultmidpunct}
{}{\mcitedefaultseppunct}\relax
\EndOfBibitem
\bibitem[Shkulipa \emph{et~al.}(2006)Shkulipa, den Otter, and
  Briels]{Shkulipa:2006aa}
S.~A. Shkulipa, W.~K. den Otter and W.~J. Briels, \emph{Phys Rev Lett}, 2006,
  \textbf{96}, 178302\relax
\mciteBstWouldAddEndPuncttrue
\mciteSetBstMidEndSepPunct{\mcitedefaultmidpunct}
{\mcitedefaultendpunct}{\mcitedefaultseppunct}\relax
\EndOfBibitem
\bibitem[Campillo \emph{et~al.}(2013)Campillo, Sens, K{\"o}ster, Pontani,
  L{\'e}vy, Bassereau, Nassoy, and Sykes]{Campillo:2013aa}
C.~Campillo, P.~Sens, D.~K{\"o}ster, L.-L. Pontani, D.~L{\'e}vy, P.~Bassereau,
  P.~Nassoy and C.~Sykes, \emph{Biophys J}, 2013, \textbf{104}, 1248--56\relax
\mciteBstWouldAddEndPuncttrue
\mciteSetBstMidEndSepPunct{\mcitedefaultmidpunct}
{\mcitedefaultendpunct}{\mcitedefaultseppunct}\relax
\EndOfBibitem
\bibitem[Bas()]{Bassereau-Mangenot}
P. Bassereau and S. Mangenot, personal communication.\relax
\mciteBstWouldAddEndPunctfalse
\mciteSetBstMidEndSepPunct{\mcitedefaultmidpunct}
{}{\mcitedefaultseppunct}\relax
\EndOfBibitem
\bibitem[Goulian \emph{et~al.}(1998)Goulian\emph{et~al.}]{GoulianBJ98}
M.~Goulian \emph{et~al.}, \emph{Biophys. J.}, 1998, \textbf{74}, 328\relax
\mciteBstWouldAddEndPuncttrue
\mciteSetBstMidEndSepPunct{\mcitedefaultmidpunct}
{\mcitedefaultendpunct}{\mcitedefaultseppunct}\relax
\EndOfBibitem
\end{mcitethebibliography}


\begin{thebibliography}{10}

\bibitem{long_fully_2015}
Long, J., Shelhamer, E., Darrell, T.:
\newblock Fully {Convolutional} {Networks} for {Semantic} {Segmentation}.
\newblock In: Proceedings of the {IEEE} {Conference} on {Computer} {Vision} and
  {Pattern} {Recognition}. (2015)  3431--3440

\bibitem{everingham_pascal_2014}
Everingham, M., Eslami, S.M.A., Gool, L.V., Williams, C.K.I., Winn, J.,
  Zisserman, A.:
\newblock The {Pascal} {Visual} {Object} {Classes} {Challenge}: {A}
  {Retrospective}.
\newblock International Journal of Computer Vision \textbf{111} (2014)  98--136

\bibitem{lin_microsoft_2014}
Lin, T.Y., Maire, M., Belongie, S., Hays, J., Perona, P., Ramanan, D., Dollár,
  P., Zitnick, C.L.:
\newblock Microsoft {COCO}: {Common} {Objects} in {Context}.
\newblock In Fleet, D., Pajdla, T., Schiele, B., Tuytelaars, T., eds.: Computer
  {Vision} – {ECCV} 2014. Number 8693 in Lecture {Notes} in {Computer}
  {Science}.
\newblock Springer International Publishing (2014)  740--755 DOI:
  10.1007/978-3-319-10602-1\_48.

\bibitem{lagrange_benchmarking_2015}
Lagrange, A., Le~Saux, B., Beaupere, A., Boulch, A., Chan-Hon-Tong, A., Herbin,
  S., Randrianarivo, H., Ferecatu, M.:
\newblock Benchmarking classification of earth-observation data: {From}
  learning explicit features to convolutional networks.
\newblock In: {IEEE} {International} {Geosciences} and {Remote} {Sensing}
  {Symposium} ({IGARSS}). (2015)  4173--4176

\bibitem{paisitkriangkrai_effective_2015}
Paisitkriangkrai, S., Sherrah, J., Janney, P., Van Den~Hengel, A.:
\newblock Effective semantic pixel labelling with convolutional networks and
  {Conditional} {Random} {Fields}.
\newblock In: Proceedings of the {IEEE} {Conference} on {Computer} {Vision} and
  {Pattern} {Recognition} {Workshops}. (2015)  36--43

\bibitem{rottensteiner_isprs_2012}
Rottensteiner, F., Sohn, G., Jung, J., Gerke, M., Baillard, C., Benitez, S.,
  Breitkopf, U.:
\newblock The {ISPRS} benchmark on urban object classification and 3d building
  reconstruction.
\newblock ISPRS Ann. Photogramm. Remote Sens. Spat. Inf. Sci \textbf{1} (2012)
  ~3

\bibitem{chen_semantic_2015}
Chen, L.C., Papandreou, G., Kokkinos, I., Murphy, K., Yuille, A.:
\newblock Semantic {Image} {Segmentation} with {Deep} {Convolutional} {Nets}
  and {Fully} {Connected} {CRFs}.
\newblock In: Proceedings of the {International} {Conference} on {Learning}
  {Representations}. (2015)

\bibitem{yu_multi-scale_2015}
Yu, F., Koltun, V.:
\newblock Multi-{Scale} {Context} {Aggregation} by {Dilated} {Convolutions}.
\newblock In: Proceedings of the {International} {Conference} on {Learning}
  {Representations}. (2015)

\bibitem{zheng_conditional_2015}
Zheng, S., Jayasumana, S., Romera-Paredes, B., Vineet, V., Su, Z., Du, D.,
  Huang, C., Torr, P.H.S.:
\newblock Conditional {Random} {Fields} as {Recurrent} {Neural} {Networks}.
\newblock In: Proceedings of the {IEEE} {International} {Conference} on
  {Computer} {Vision}. (2015)  1529--1537

\bibitem{arnab_higher_2015}
Arnab, A., Jayasumana, S., Zheng, S., Torr, P.:
\newblock Higher {Order} {Conditional} {Random} {Fields} in {Deep} {Neural}
  {Networks}.
\newblock arXiv:1511.08119 [cs] (2015) arXiv: 1511.08119.

\bibitem{he_deep_2016}
He, K., Zhang, X., Ren, S., Sun, J.:
\newblock Deep {Residual} {Learning} for {Image} {Recognition}.
\newblock In: Proceedings of the {IEEE} {Conference} on {Computer} {Vision} and
  {Pattern} {Recognition}. (2016)

\bibitem{wu_high-performance_2016}
Wu, Z., Shen, C., Van Den~Hengel, A.:
\newblock High-performance {Semantic} {Segmentation} {Using} {Very} {Deep}
  {Fully} {Convolutional} {Networks}.
\newblock arXiv:1604.04339 [cs] (2016) arXiv: 1604.04339.

\bibitem{yan_combining_2016}
Yan, Z., Zhang, H., Jia, Y., Breuel, T., Yu, Y.:
\newblock Combining the {Best} of {Convolutional} {Layers} and {Recurrent}
  {Layers}: {A} {Hybrid} {Network} for {Semantic} {Segmentation}.
\newblock arXiv:1603.04871 [cs] (2016) arXiv: 1603.04871.

\bibitem{zhao_stacked_2015}
Zhao, J., Mathieu, M., Goroshin, R., LeCun, Y.:
\newblock Stacked {What}-{Where} {Auto}-encoders.
\newblock In: Proceedings of the {International} {Conference} on {Learning}
  {Representations}. (2015)

\bibitem{noh_learning_2015}
Noh, H., Hong, S., Han, B.:
\newblock Learning {Deconvolution} {Network} for {Semantic} {Segmentation}.
\newblock In: Proceedings of the {IEEE} {Conference} on {Computer} {Vision} and
  {Pattern} {Recognition}. (2015)  1520--1528

\bibitem{badrinarayanan_segnet:_2015}
Badrinarayanan, V., Kendall, A., Cipolla, R.:
\newblock {SegNet}: {A} {Deep} {Convolutional} {Encoder}-{Decoder}
  {Architecture} for {Image} {Segmentation}.
\newblock arXiv preprint arXiv:1511.00561 (2015)

\bibitem{mnih_learning_2010}
Mnih, V., Hinton, G.E.:
\newblock Learning to {Detect} {Roads} in {High}-{Resolution} {Aerial}
  {Images}.
\newblock In Daniilidis, K., Maragos, P., Paragios, N., eds.: Computer {Vision}
  – {ECCV} 2010. Number 6316 in Lecture {Notes} in {Computer} {Science}.
\newblock Springer Berlin Heidelberg (2010)  210--223

\bibitem{penatti_deep_2015}
Penatti, O., Nogueira, K., Dos~Santos, J.:
\newblock Do deep features generalize from everyday objects to remote sensing
  and aerial scenes domains?
\newblock In: Proceedings of the {IEEE} {Conference} on {Computer} {Vision} and
  {Pattern} {Recognition} {Workshops}. (2015)  44--51

\bibitem{campos-taberner_processing_2016}
Campos-Taberner, M., Romero-Soriano, A., Gatta, C., Camps-Valls, G., Lagrange,
  A., Le~Saux, B., Beaupère, A., Boulch, A., Chan-Hon-Tong, A., Herbin, S.,
  Randrianarivo, H., Ferecatu, M., Shimoni, M., Moser, G., Tuia, D.:
\newblock Processing of {Extremely} {High}-{Resolution} {LiDAR} and {RGB}
  {Data}: {Outcome} of the 2015 {IEEE} {GRSS} {Data} {Fusion} {Contest} {Part}
  {A}: 2-{D} {Contest}.
\newblock IEEE Journal of Selected Topics in Applied Earth Observations and
  Remote Sensing \textbf{PP} (2016)  1--13

\bibitem{nogueira_towards_2016}
Nogueira, K., Penatti, O.A.B., Dos~Santos, J.A.:
\newblock Towards {Better} {Exploiting} {Convolutional} {Neural} {Networks} for
  {Remote} {Sensing} {Scene} {Classification}.
\newblock arXiv:1602.01517 [cs] (2016) arXiv: 1602.01517.

\bibitem{zhao_learning_2016}
Zhao, W., Du, S.:
\newblock Learning multiscale and deep representations for classifying remotely
  sensed imagery.
\newblock ISPRS Journal of Photogrammetry and Remote Sensing \textbf{113}
  (2016)  155--165

\bibitem{marmanis_semantic_2016}
Marmanis, D., Wegner, J.D., Galliani, S., Schindler, K., Datcu, M., Stilla, U.:
\newblock Semantic {Segmentation} of {Aerial} {Images} with an {Ensemble} of
  {CNNs}.
\newblock ISPRS Annals of Photogrammetry, Remote Sensing and Spatial
  Information Sciences \textbf{3} (2016)  473--480

\bibitem{gerke_use_2015}
Gerke, M.:
\newblock Use of the {Stair} {Vision} {Library} within the {ISPRS} 2d
  {Semantic} {Labeling} {Benchmark} ({Vaihingen}).
\newblock Technical report, International Institute for Geo-Information Science
  and Earth Observation (2015)

\bibitem{chatfield_return_2014}
Chatfield, K., Simonyan, K., Vedaldi, A., Zisserman, A.:
\newblock Return of the {Devil} in the {Details}: {Delving} {Deep} into
  {Convolutional} {Nets}.
\newblock In: Proceedings of the {British} {Machine} {Vision} {Conference},
  British Machine Vision Association (2014)  6.1--6.12

\bibitem{simonyan_very_2014}
Simonyan, K., Zisserman, A.:
\newblock Very {Deep} {Convolutional} {Networks} for {Large}-{Scale} {Image}
  {Recognition}.
\newblock arXiv:1409.1556 [cs] (2014) arXiv: 1409.1556.

\bibitem{ioffe_batch_2015}
Ioffe, S., Szegedy, C.:
\newblock Batch {Normalization}: {Accelerating} {Deep} {Network} {Training} by
  {Reducing} {Internal} {Covariate} {Shift}.
\newblock In: Proceedings of the 32nd {International} {Conference} on {Machine}
  {Learning}. (2015)  448--456

\bibitem{he_delving_2015}
He, K., Zhang, X., Ren, S., Sun, J.:
\newblock Delving {Deep} into {Rectifiers}: {Surpassing} {Human}-{Level}
  {Performance} on {ImageNet} {Classification}.
\newblock In: Proceedings of the {IEEE} {International} {Conference} on
  {Computer} {Vision}. (2015)  1026--1034

\bibitem{clevert_fast_2015}
Clevert, D.A., Unterthiner, T., Hochreiter, S.:
\newblock Fast and {Accurate} {Deep} {Network} {Learning} by {Exponential}
  {Linear} {Units} ({ELUs}).
\newblock In: Proceedings of the {International} {Conference} on {Learning}
  {Representations}. (2015)

\bibitem{marmanis_deep_2016}
Marmanis, D., Datcu, M., Esch, T., Stilla, U.:
\newblock Deep {Learning} {Earth} {Observation} {Classification} {Using}
  {ImageNet} {Pretrained} {Networks}.
\newblock IEEE Geoscience and Remote Sensing Letters \textbf{13} (2016)
  105--109

\bibitem{szegedy_going_2015}
Szegedy, C., Liu, W., Jia, Y., Sermanet, P., Reed, S., Anguelov, D., Erhan, D.,
  Vanhoucke, V., Rabinovich, A.:
\newblock Going deeper with convolutions.
\newblock In: Proceedings of the {IEEE} {Conference} on {Computer} {Vision} and
  {Pattern} {Recognition}. (2015)  1--9

\bibitem{liao_video_2015}
Liao, R., Tao, X., Li, R., Ma, Z., Jia, J.:
\newblock Video {Super}-{Resolution} via {Deep} {Draft}-{Ensemble} {Learning}.
\newblock In: Proceedings of the {IEEE} {International} {Conference} on
  {Computer} {Vision}. (2015)  531--539

\bibitem{liao_competitive_2015}
Liao, Z., Carneiro, G.:
\newblock Competitive {Multi}-scale {Convolution}.
\newblock arXiv:1511.05635 [cs] (2015) arXiv: 1511.05635.

\bibitem{ngiam_multimodal_2011}
Ngiam, J., Khosla, A., Kim, M., Nam, J., Lee, H., Ng, A.Y.:
\newblock Multimodal deep learning.
\newblock In: Proceedings of the 28th international conference on machine
  learning ({ICML}-11). (2011)  689--696

\bibitem{eitel_multimodal_2015}
Eitel, A., Springenberg, J.T., Spinello, L., Riedmiller, M., Burgard, W.:
\newblock Multimodal deep learning for robust {RGB}-{D} object recognition.
\newblock In: Proceedings of the {International} {Conference} on {Intelligent}
  {Robots} and {Systems}, IEEE (2015)  681--687

\bibitem{quang_efficient_2015}
Quang, N.T., Thuy, N.T., Sang, D.V., Binh, H.T.T.:
\newblock An {Efficient} {Framework} for {Pixel}-wise {Building} {Segmentation}
  from {Aerial} {Images}.
\newblock In: Proceedings of the {Sixth} {International} {Symposium} on
  {Information} and {Communication} {Technology}, ACM (2015) ~43

\bibitem{boulch_dag_2015}
Boulch, A.:
\newblock {DAG} of convolutional networks for semantic labeling.
\newblock Technical report, Office national d'études et de recherches
  aérospatiales (2015)

\bibitem{lin_efficient_2015}
Lin, G., Shen, C., Van Den~Hengel, A., Reid, I.:
\newblock Efficient piecewise training of deep structured models for semantic
  segmentation.
\newblock In: Proceedings of the {IEEE} {Conference} on {Computer} {Vision} and
  {Pattern} {Recognition}. (2015)

\bibitem{yosinski_how_2014}
Yosinski, J., Clune, J., Bengio, Y., Lipson, H.:
\newblock How transferable are features in deep neural networks?
\newblock In: Advances in {Neural} {Information} {Processing} {Systems}. (2014)
   3320--3328

\bibitem{cramer_dgpf_2010}
Cramer, M.:
\newblock The {DGPF} test on digital aerial camera evaluation – overview and
  test design.
\newblock Photogrammetrie – Fernerkundung – Geoinformation \textbf{2}
  (2010)  73--82

\end{thebibliography}



\end{document}